\documentclass[10pt,twocolumn,letterpaper]{article}

\usepackage{cvpr}              
\usepackage{amsmath}
\usepackage{bm}

\usepackage[dvipsnames]{xcolor}

\definecolor{cvprblue}{rgb}{0.21,0.49,0.74}
\usepackage[pagebackref,breaklinks,colorlinks,citecolor=cvprblue]{hyperref}

\def\paperID{8390} 
\def\confName{CVPR}
\def\confYear{2025}

\title{Satellite to GroundScape - Large-scale Consistent Ground View Generation from Satellite Views}

\author{Ningli Xu\\
The Ohio State University\\
{\tt\small xu.3961@buckeyemail.osu.edu}
\and
Rongjun Qin\\
The Ohio State University\\
{\tt\small qin.324@osu.edu}
}

\begin{document}
\maketitle
\let\thefootnote\relax\footnotetext{Pre-print version: to be published in CVPR 2025}
\begin{abstract}
Generating consistent ground-view images from satellite imagery is challenging, primarily due to the large discrepancies in viewing angles and resolution between satellite and ground-level domains. Previous efforts mainly concentrated on single-view generation, often resulting in inconsistencies across neighboring ground views. In this work, we propose a novel cross-view synthesis approach designed to overcome these challenges by ensuring consistency across ground-view images generated from satellite views. Our method, based on a fixed latent diffusion model, introduces two conditioning modules: satellite-guided denoising, which extracts high-level scene layout to guide the denoising process, and satellite-temporal denoising, which captures camera motion to maintain consistency across multiple generated views. We further contribute a large-scale satellite-ground dataset containing over 100,000 perspective pairs to facilitate extensive ground scene or video generation. Experimental results demonstrate that our approach outperforms existing methods on perceptual and temporal metrics, achieving high photorealism and consistency in multi-view outputs. The project page is at \url{https://gdaosu.github.io/sat2groundscape}. 
\end{abstract}    
\section{Introduction}
\label{sec:intro}
\begin{figure}[t]
    \centering
    \includegraphics[width=\linewidth]{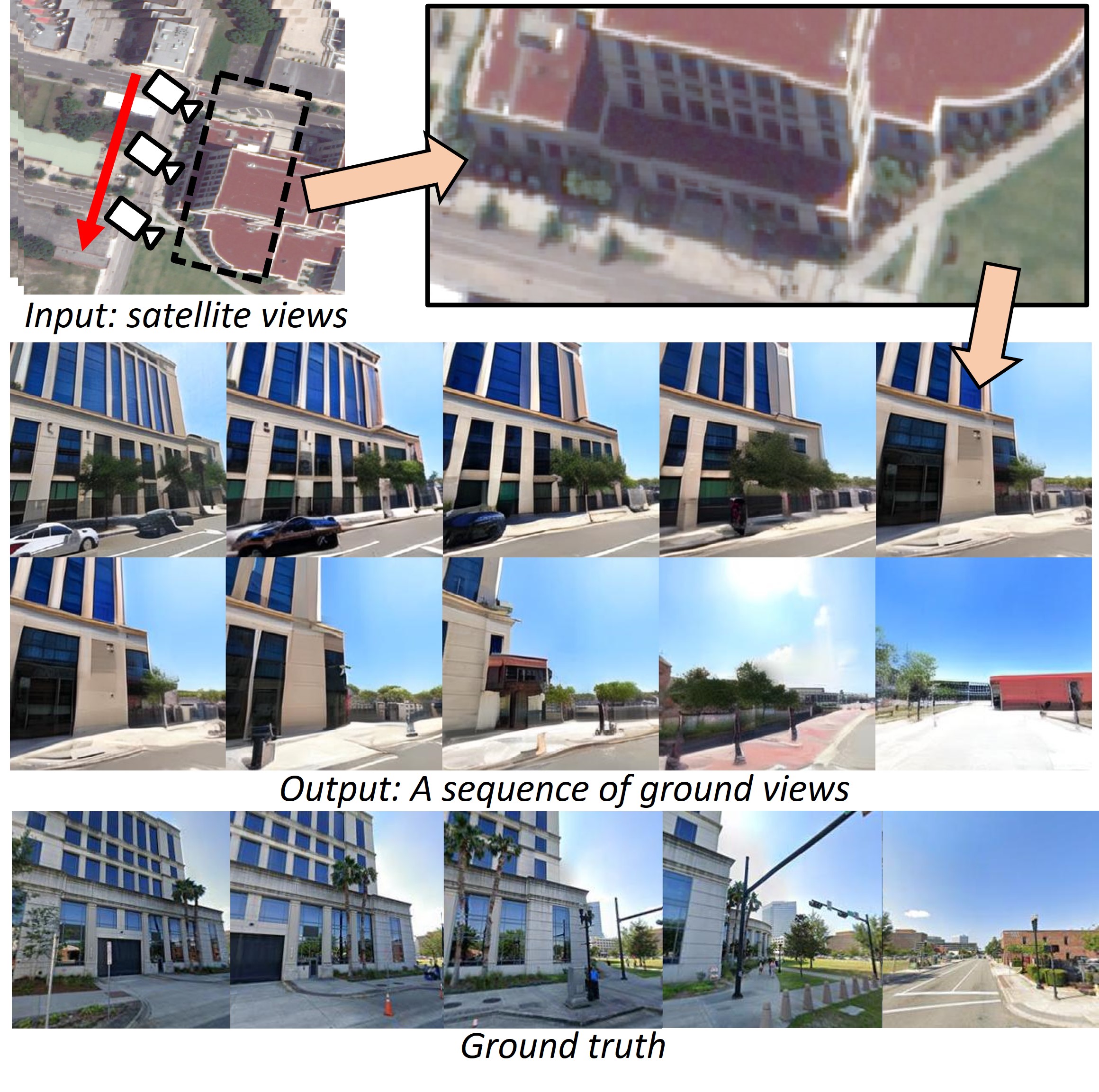}
    \caption{\textbf{Ground views generated by Sat2GroundScape.} Using satellite views as input, Sat2GroundScape generates a sequence of ground views that exhibit photorealistic quality and maintain consistent ground appearances across different perspectives.}
    \label{fig:teaser}
\end{figure}
The growing availability of satellite imagery has unlocked new opportunities for generating realistic ground scene representations from top-down satellite views, a process known as cross-view synthesis. This capability holds significant potential for applications like immersive 3D gaming and large-scale urban modeling, providing a richer medium for visualizing environments from the ground level \cite{Li_2024_CVPR, lu2020geometry, xu2024multi}. However, the task of generating consistent ground-view images across multiple perspectives presents additional challenges, complicating an already intricate problem.

The key challenges stem from establishing a reliable and stable mapping between the satellite and ground domains. The generated ground views must not only adhere to the scene layouts indicated by the satellite data but also maintain consistency across multiple ground perspectives. The substantial gap between satellite and ground imagery (marked by nearly 90-degree differences in viewing angles and a resolution disparity of almost ten times \cite{xu2024geospecific}) makes it particularly challenging to establish such a stable connection between the satellite and ground domains.

Although recent approaches \cite{ren2021cascaded, wu2022cross, lu2020geometry, xu2024geospecific} have made significant progress in addressing these challenges, they typically achieve impressive results in single-view synthesis. However, these results often lack consistency in appearance when applied to multi-view synthesis. This issue arises because generative models, such as GANs and diffusion models, are employed for conditional image generation, which introduces randomness, especially in regions where no guidance is available (e.g., shadowed or textureless areas). Some methods \cite{lu2020geometry, regmi2018cross, ren2021cascaded} estimate ground layouts in a semantic format and use cGAN-based techniques \cite{isola2017image, choi2018stargan} to generate ground views conditioned on these semantic layouts. GVG \cite{xu2024geospecific} proposed a diffusion-based approach that generates ground views by conditioning on the projected satellite texture at the ground level, framing the problem as an image super-resolution task. However, all these methods fail to maintain consistency across neighboring ground views. Specifically, facade details critical for urban environments are either lost or inconsistently rendered across views, limiting the practical applicability of the synthesized images in realistic scenarios.

In this work, we propose a novel approach for satellite-to-ground view synthesis that ensures consistency across the generated ground views, as shown in \cref{fig:teaser}. Building on the LDM \cite{rombach2022high}, we introduce a satellite-guided denoising process to bridge the significant domain gap between satellite and ground imagery. This process enables the pre-trained LDM to produce ground views that preserve the same scene layouts as the satellite inputs.To generate multiple consistent ground views, we further propose a satellite-temporal denoising process that captures camera motion from the satellite conditions.  Additionally, we present Sat2GroundScape, a large-scale satellite-ground dataset, to support extensive ground scene and video generation from satellite imagery. Experimental results demonstrate that our method surpasses all baselines on both perceptual and temporal metrics. The key contributions of this work are as follows:
\begin{itemize}
    \item We introduce a satellite-to-ground synthesis framework that ensures consistency across multiple generated ground views by employing satellite-guided denoising and satellite-temporal denoising processes, creating a stable and coherent link between satellite and ground domains.
    \item We introduce Sat2GroundScape, a dataset with 25,000+ panoramic and 100,000+ perspective satellite-ground image pairs for ground scene generation.
    \item Our method outperforms SOTA in perceptual and temporal metrics, achieving high photorealism and consistency. 
\end{itemize}


\section{Related work}
\label{sec:related_work}
\subsection{Cross-view synthesis}
Cross-view synthesis tackles the problem of generating novel viewpoints of objects or scenes from substantially different perspectives. A representative task in this area involves synthesizing ground-level views based on top-down satellite images. Due to the substantial difference in viewport and resolution, novel view synthesis methods, such as NeRF \cite{mildenhall2021nerf} or Gaussian Splatting \cite{kerbl20233d}, are ineffective for cross-view synthesis tasks \cite{gao2024skyeyes}. Current approaches frequently utilize generative models, such as GANs \cite{zhu2017unpaired} or LDMs \cite{rombach2022high,zhang2023adding}, to bridge these differences, conditioning generation on the satellite view or high-level features extracted from it.  Models like X-fork \cite{regmi2018cross} and PanoGAN \cite{wu2022cross} apply cycle-GAN \cite{isola2017image} to directly predict both ground-level views and corresponding semantic representations from top-down images. Sat2Ground \cite{lu2020geometry} further builds on these methods by integrating geometric consistency, estimating a height map from satellite images to transform viewpoints and predict ground-view semantics and appearance. Sat2Density \cite{qian2023sat2density} extends this by predicting top-view density without depth supervision, using the relationship between satellite and ground views along with neural rendering techniques to enhance synthesis quality. Additionally, GVG \cite{xu2024geospecific} shows that incorporating weak facade information from satellite images significantly improves ground-view generation, employing a diffusion-based model conditioned on satellite textures and edge maps. Nevertheless, these models largely focus on generating single ground views, lacking spatial consistency across neighboring views. InfiniCity \cite{lin2023infinicity} addresses this by introducing a 3D voxel grid that captures both satellite geometry and textures, generating ground views with a GAN-based neural rendering module using ray sampling within the voxel grid.  Sat2Scene \cite{Li_2024_CVPR} builds on this by employing a diffusion-based 3D sparse representation to improve synthesis. Despite these advances, limitations persist: the quality of generated ground views is restricted by voxel resolution, and scene-specific training requirements limit scalability across diverse outdoor environments.

\begin{figure*}
    \centering
    \includegraphics[width=\textwidth]{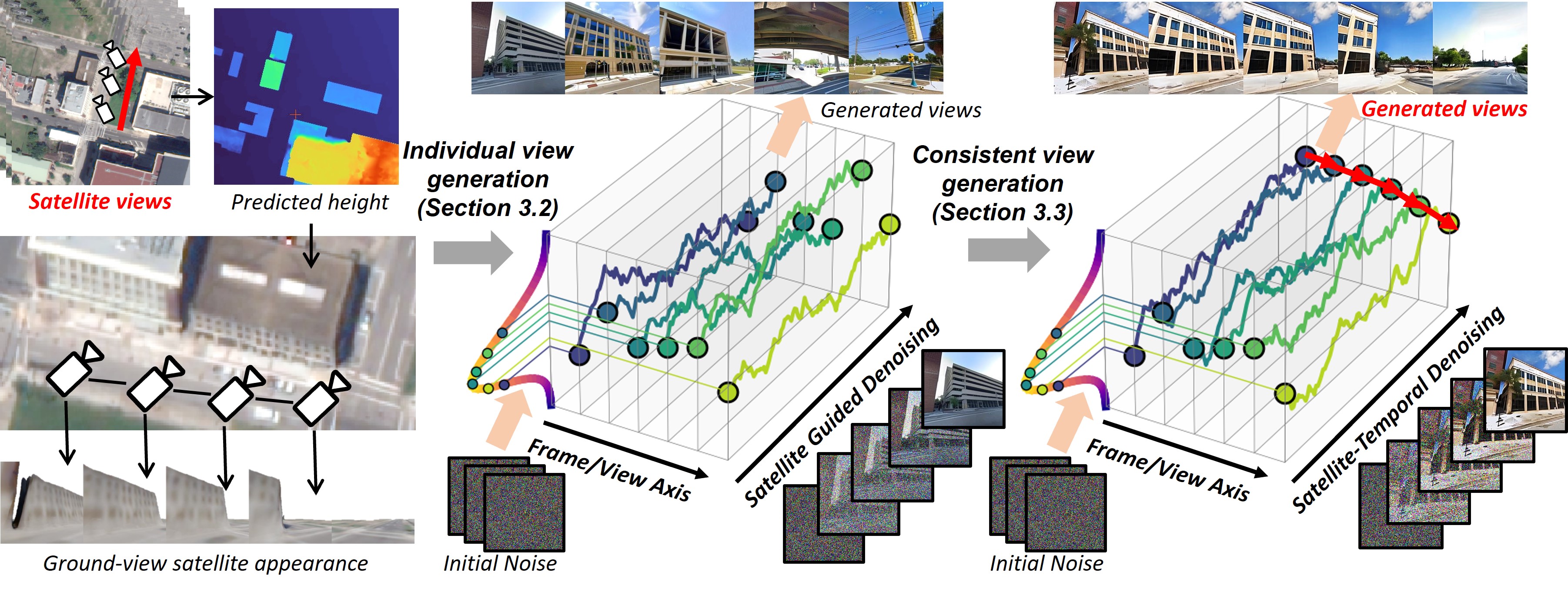}
    \caption{\textbf{Overview pipeline of Sat2GroundScape}. The satellite appearance is initially projected onto the ground level based on the estimated satellite geometry. \textbf{Satellite-Guided Denoising} is then introduced to guide the latent diffusion model (LDM) in generating individual ground views that preserve the original scene layouts. \textbf{Satellite-Temporal Denoising} is proposed to further ensure consistency across multiple generated views. Input/output are marked as red.}
    \label{fig:pipeline}
\end{figure*}

\subsection{Diffusion-based view generation} 
Diffusion models employ an iterative refinement mechanism, progressively denoising samples initialized from a normal distribution, and have emerged as leading frameworks for view generation. Since the introduction of DDPM \cite{ho2020denoising}, diffusion models have demonstrated superior stability and quality over GANs, achieving SOTA results. Subsequent advancements, such as DDIM \cite{song2020denoising} and the model proposed by \cite{nichol2021improved}, improve sampling efficiency while preserving generation quality and optimizing training schedules. LDMs \cite{rombach2022high} further enhance stability by operating within a compressed latent space, which significantly reduces computational and memory requirements. ControlNet \cite{zhang2023adding} showed that diffusion-based image generation can be flexibly conditioned on various inputs (such as edge maps, depth, layouts, and human poses) by encoding these conditions as latent residues within each U-Net block. Building on this approach, recent studies have adapted similar conditioning techniques for ground-view generation. For example, MagicDrive \cite{gao2023magicdrive} supports conditional street-view generation with extensive 3D geometric controls, including camera poses, 3D bounding boxes, and bird eye view maps. Similarly, Streetscapes \cite{deng2024streetscapes} integrates semantic, depth, and disparity information as conditioning inputs.

\subsection{Consistent view generation}
The randomness in the denoising process of diffusion models presents challenges for achieving multi-view consistency when generating multiple images of the same scene, a critical requirement for applications in scene or video generation. While research in this area is limited, some approaches have begun exploring strategies to address consistency.

\textbf{Text-to-video generation} methods leverage pre-trained text-to-image models \cite{rombach2022high} by incorporating temporal mixing layers into their architectures in various ways \cite{blattmann2023stable, peng2024controlnext, ma2024follow, wu2023tune, zhang2023controlvideo}.  For example, ControlVideo \cite{zhang2023controlvideo} inflates each convolution and attention layer in the UNet architecture into the temporal dimension, enabling the pre-trained model to produce consistent multi-view outputs without additional parameters or retraining. StableVideoDiffusion \cite{blattmann2023stable} and related methods \cite{peng2024controlnext, ma2024follow, wu2023tune} add temporal convolution and attention layers after each spatial layer, allowing a text-to-image model fine-tuned on video datasets. MVDiffusion \cite{MVDiffusion} further enhances multi-view consistency by incorporating correspondence-aware attention layers into each U-Net block to capture inter-view relationships. However, these approaches generally support only a fixed number of views, limiting scalability for larger scenes or extended video sequences. 

\textbf{3D-aware view generation} methods aim to achieve multi-view generation by respecting scene geometry. InfiniCity \cite{lin2023infinicity}, Sat2vid \cite{li2021sat2vid}, and Sat2Scene \cite{Li_2024_CVPR} frame multi-view generation as a scene appearance estimation problem, where scene geometry (e.g., 3D voxel grids or point clouds) is predefined or predicted, and appearance attributes are learned as parameters associated with these primitives.  Their multi-view consistency is maintained through neural rendering; however, this approach requires substantial computational and memory resources to represent a complete scene and involves per-scene training. Alternatively, methods such as SceneScape \cite{Scenescape} and Streetscapes \cite{deng2024streetscapes} treat multi-view generation as an autoregressive process, where an initial view is generated using standard text-to-image techniques, and subsequent views are conditioned on previous views to maintain consistency. This approach relies on scene warping for consistency but is sensitive to precise scene geometry to prevent warping distortions. In our satellite-to-ground setting, where low-resolution satellite data is used with significant spatial uncertainty, the warping process tends to introduce distortions across views and thus accumulate the distortion and artifacts, leading to generating poor-quality views after several iterations. 

\section{Sat2GroundScape}
We propose a novel pipeline for generating multiple ground views from a set of satellite images, as illustrated in \cref{fig:pipeline}. The process begins with estimating the scene geometry from satellite views, which enables the projection of satellite-based appearance onto the ground level, followed by satellite-guided denoising to estimate an initial ground view (\cref{sec:individual}). Subsequently, consistent ground view generation is attained through a satellite-temporal denoising process (\cref{sec:temporal}). Additionally, Furthermore, we introduce a large-scale satellite-ground dataset designed to support large-scale ground scene or video generation (\cref{sec:dataset}).

\subsection{Background on latent denoising process}
In image generation, the objective of diffusion models is to sample images from an underlying data distribution \(p(\boldsymbol{x})\). A typical denoising process is to iteratively denoise samples from random noise into samples from the data distribution \(\boldsymbol{x}_0 \sim p(\boldsymbol{x})\). LDM \cite{rombach2022high} have demonstrated that conducting this denoising process in a latent feature space significantly enhances stability and efficiency. Given a randomly initialized noisy image \(\boldsymbol{x}_T\) in pixel space which is first encoded as a latent feature \(\boldsymbol{z}_T=\mathcal{E}(\boldsymbol{x}_T)\), a LDM iteratively denoises \(\boldsymbol{z}_T\) to obtain \(\boldsymbol{z}_0\) over a series of \(T\) denoising steps. The final denoised feature \(\boldsymbol{z}_0\) is then decoded back to pixel space as \(\boldsymbol{x}_0=\mathcal{D}(\boldsymbol{z}_0)\). Here, \(\mathcal{E}\) and \(\mathcal{D}\) represent pre-trained encoders and decoders that map between pixel space and latent space \cite{kingma2013auto}. This latent denoising process is formalized as follows:
\begin{equation}
    \boldsymbol{z}_{t-1}=DDIM (\boldsymbol{z}_t, \epsilon_\theta(\boldsymbol{z}_t,t),t)
\end{equation}
where \(\epsilon\) represents a neural network with learned parameters \(\theta\) that predicts the noise component; The DDIM denoiser \cite{song2020denoising} is then employed to compute \(z_{t-1}\) from this prediction.

\begin{figure}[t]
    \centering
    \includegraphics[width=\linewidth]{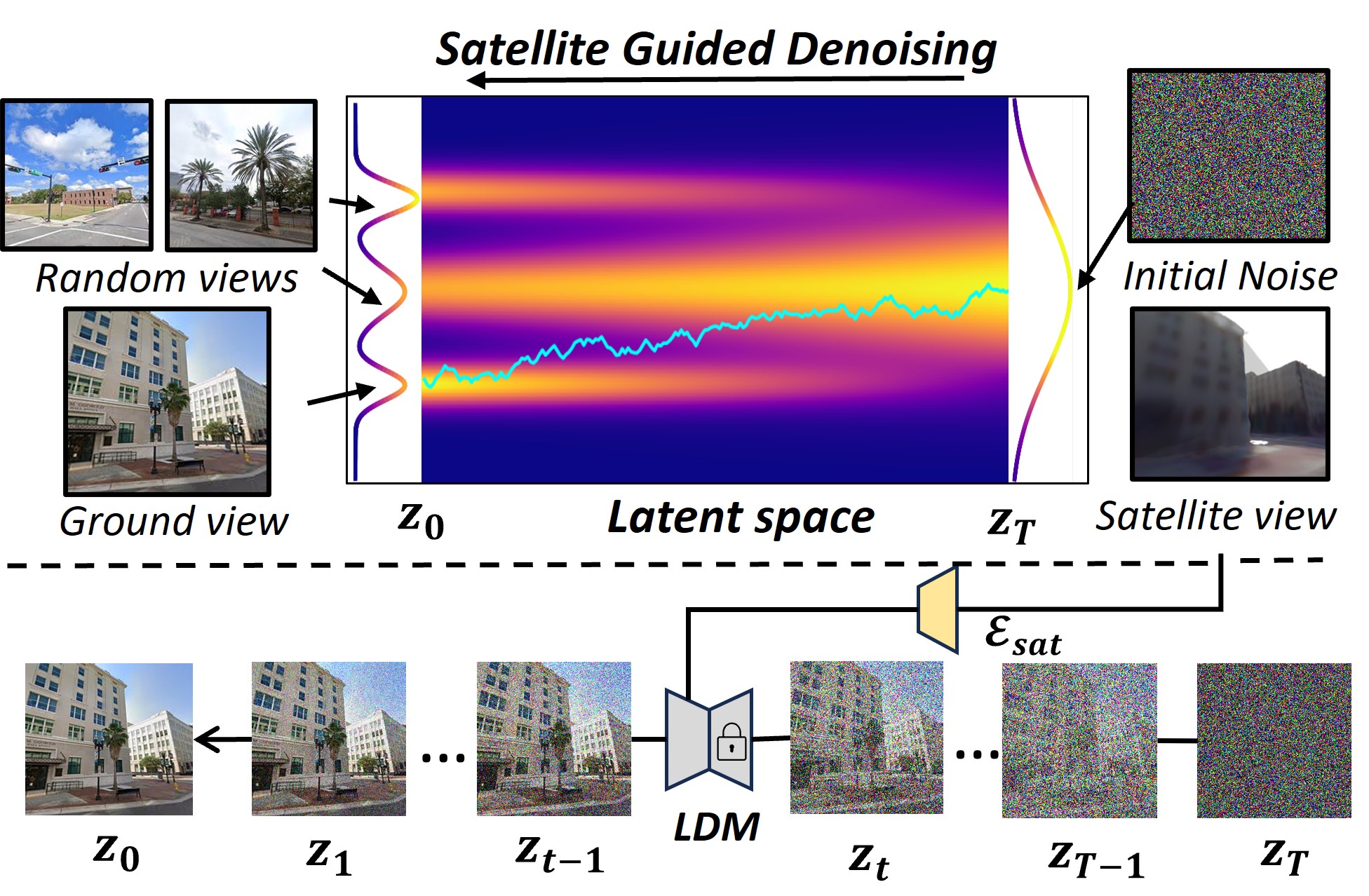}
    \caption{\textbf{Satellite-Guided Denoising.} Conditioning on a given satellite view, a random noisy latent feature \(\boldsymbol{z}_T\) is iteratively denoised to finally become the corresponding ground view latent feature \(\boldsymbol{z}_0\) instead of other randomly generated ground views. We extract the high-level satellite features and guide the standard LDM to perform denoising. Note that \(\boldsymbol{z}_i\) are in latent spaces, we illustrate these latent features with corresponding images in pixel space.}
    \label{fig:sat_cond}
\end{figure}

\subsection{Individual ground view generation}
\label{sec:individual}
Our approach initializes the 3D scene in a format optimized to retain as much of the original satellite data as possible, supporting both camera control and ground-view generation. Similar to GVG \cite{xu2024geospecific}, we represent the scene as a unified triangle mesh, which offers a dense representation of scene geometry, appearance, and visibility, computed through traditional multi-view stereo methods \cite{hirschmuller2005accurate}. The model appearance is derived using texture mapping \cite{ling2023large}. Given pre-defined ground cameras \(\{\boldsymbol{C}^{i}\}\), we render the ground-view satellite appearance \(\{\boldsymbol{I}_g^{i}\}\) from the satellite mesh. This rendering technique projects satellite data from 3D space into screen space, providing direct control over ground camera poses and satellite data.

\begin{figure}[t]
    \centering
    \includegraphics[width=\linewidth]{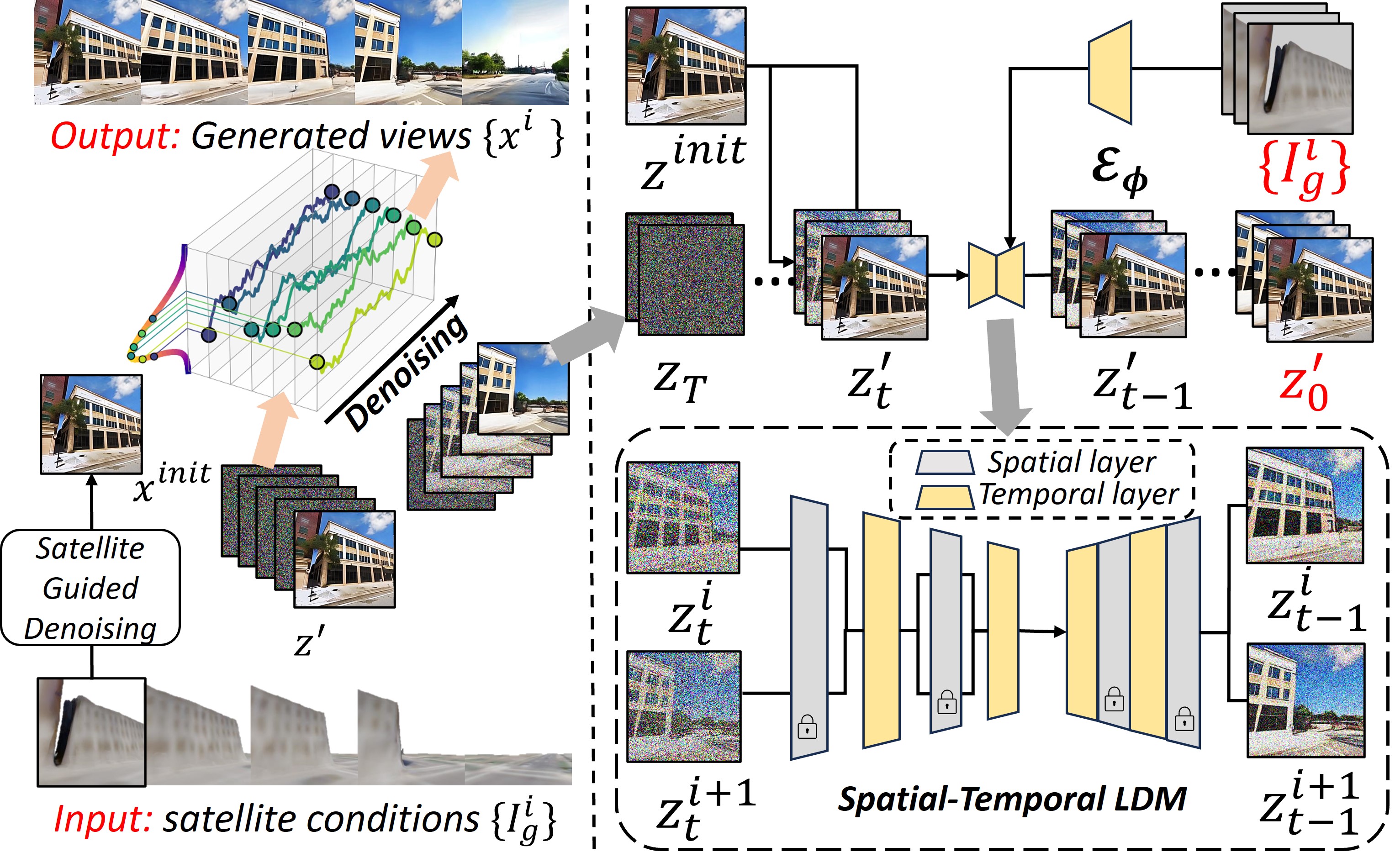}
    \caption{\textbf{Satellite-Temporal Denoising} takes a sequence of ground-view satellite appearance \(\{\boldsymbol{I}_g^i\}\) as input and generates the consistent ground views \(\{\boldsymbol{x}^i\}\). It first generates the initial ground view \(x^{init}\) and concatenates it to the initial noise as the input to the spatial-temporal LDM. Additionally, \(\{\boldsymbol{I}_g^i\}\) are encoded as camera motion features to guide the denoising process. Red variables are the input/output for our method.}
    \label{fig:temporal}
\end{figure}

\noindent\textbf{Satellite-guided denosing}. 
The satellite appearance primarily provides high-level ground layout information with limited texture detail. Our framework builds on a pre-trained LDM \cite{rombach2022high}, denoted by \(\epsilon_\theta\), and integrates additional modules to guide the denoising process. Inspired by previous works \cite{zhang2023adding, xu2024geospecific}, we adopt a UNet architecture, \(\mathcal{E}_{sat}\), to extract the high-level features from the satellite appearance \(\boldsymbol{I}_g\), resulting in \(\boldsymbol{c}_{sat}=\mathcal{E}_{sat}(\boldsymbol{I}_g)\). This extracted feature \(\boldsymbol{c}_{sat}\) then guides the latent denoising process, facilitating the generation of high-fidelity ground-view images that maintain similar layouts and appearances to the satellite input, as illustrated in \cref{fig:sat_cond}. We define this as the satellite-guided denoising process, represented as 
\begin{equation}
    \boldsymbol{z}_{t-1}=DDIM (\boldsymbol{z}_t, \epsilon_\theta(\boldsymbol{z}_t,\boldsymbol{c}_{sat},t),t)
\end{equation}
In contrast to GVG \cite{xu2024geospecific}, we maintain fixed parameters for the standard LDM, \(\epsilon_{\theta}\), throughout both training and inference stages, relying solely on satellite appearance without additional information. Such guidance is achieved by incorporating the extracted feature \(\boldsymbol{c}_{sat}\) as residues in each layer of the LDM; see supplementary material for detailed network architecture. 

\begin{figure*}
    \centering
    \includegraphics[width=\textwidth]{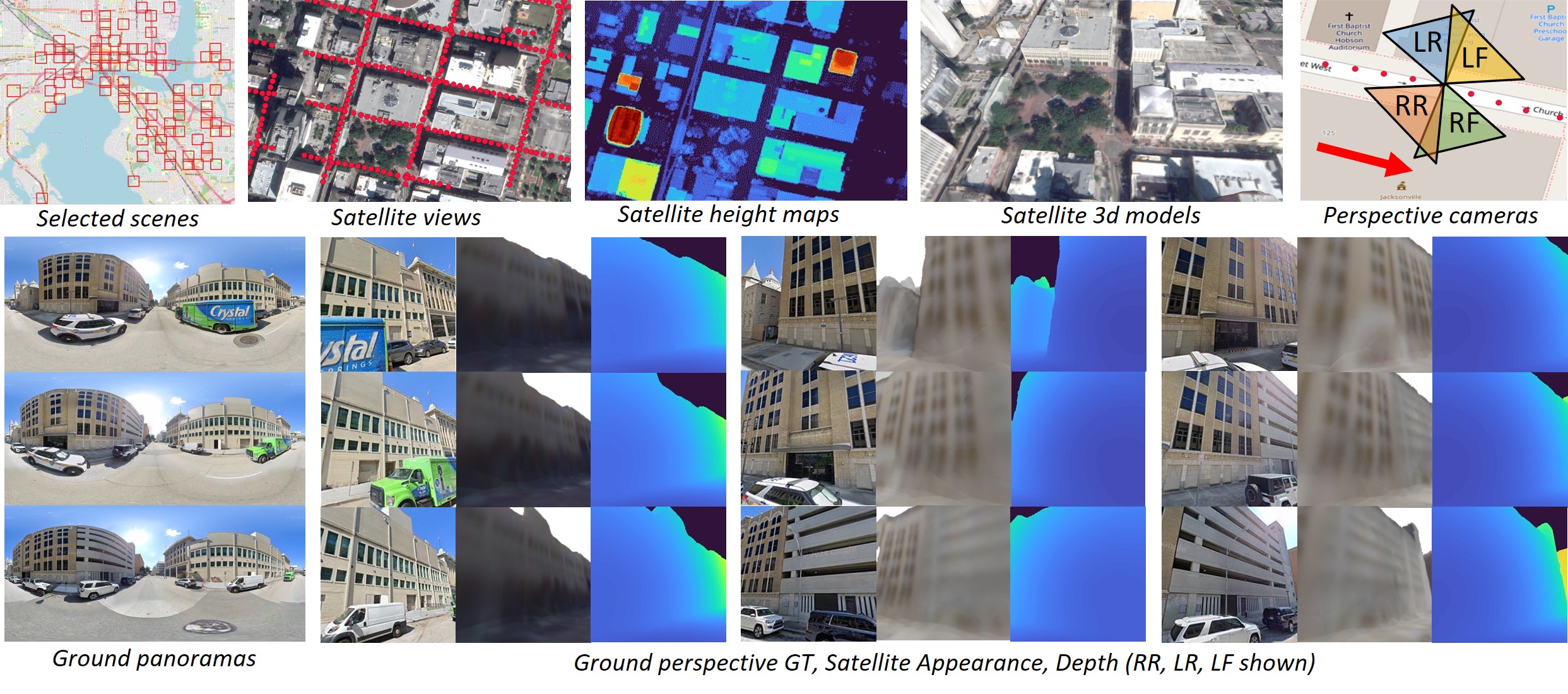}
    \caption{\textbf{Sat2GroundScape dataset.} Our dataset provides accurately aligned satellite and ground data, containing appearance, depth, and camera pose information, in both panoramic (over 25,000 pairs) and perspective formats (over 100,000 pairs). Each ground panorama is associated with four perspective views, labeled as "LF, LR, RF, RR" (left forward, left rear, right forward, and right rear). Furthermore, we include a dense ground collection (marked as "red dots") with intervals of 3 to 10 meters between points, supporting large-scale scene and video generation tasks.}
    \label{fig:dataset}
\end{figure*}

\subsection{Consistent ground views generation}
\label{sec:temporal}
After bridging the satellite-ground domain gap with the satellite-guided denoising process, we enhance consistency across multiple views in the ground domain to ensure stable, coherent ground scapes from satellite views. We introduce a \textbf{satellite-temporal denoising} process that generates consistent views conditioned on both satellite data and the previously generated ground view, as shown in \cref{fig:temporal}

For a sequence of satellite appearances \(\{\boldsymbol{I}_g^i\}\) in ground-view format, we first generate an initial ground view \(\boldsymbol{x}^{init}\) by applying our satellite-guided denoising on the first condition \(\boldsymbol{I}_g^0\). Our satellite-temporal denoising process is conditioned on \(\boldsymbol{x}^{init}\) and \(\{\boldsymbol{I}_g^i\}\). Since they are from two different domains, a critical aspect of this denoising process is designing an effective conditioning mechanism that captures information from both sources.

The denoising process begins with a random noisy latent feature \(\boldsymbol{z} \in \mathbb{R}^{T \times C \times H \times W}\), where \(T\) is the number of views to generate, and \(C,H,W\) are the spatial dimension of each view in latent space. \(\boldsymbol{x}^{init}\) can directly serve as a strong condition such that the appearance of generated views should maximally respect the \(\boldsymbol{x}^{init}\). The latent feature of \(\boldsymbol{x}^{init}\) is duplicated \(T\) times, yielding \(\boldsymbol{z}^{init} \in \mathbb{R}^{T \times C \times H \times W}\), and concatenated with \(\boldsymbol{z}\) to form the temporal-aware latent feature \(\boldsymbol{z}'\):
\begin{equation}
    \boldsymbol{z}'=[\boldsymbol{z}^{init},\boldsymbol{z}]
\end{equation}
where \(\boldsymbol{z}' \in \mathbb{R}^{T \times 2C \times H \times W}\). To handle this temporal-aware feature, we extend the pre-trained LDM model to a temporal-spatial architecture, termed temporal-spatial LDM \(\epsilon_{\phi}\), which takes \(\boldsymbol{z}'\) as input. This model iteratively estimates the noises \(\epsilon' \in \mathbb{R}^{T \times C \times H \times W}\) and denoises \(\boldsymbol{z}\). Specifically, a temporal layer is added after each spatial layer in the LDM, allowing the spatial layers to process \(\boldsymbol{z}'\) as \(T\) independent images while the temporal layers interpret \(\boldsymbol{z}'\) as a single feature for inter-views learning.  

In addition to \(\boldsymbol{x}^{init}\), satellite conditions provide camera motion and high-level layout cues. A ResNet architecture, \(\mathcal{E}_{\phi}\), is employed to extract the high-level camera motion features \(\boldsymbol{c}_{\phi}=\mathcal{E}_{\phi}(\boldsymbol{I}_g)\). Similar to \(\boldsymbol{c}_{sat}\) mentioned in \cref{sec:individual}, \(\boldsymbol{c}_{\phi}\) serves as residuals at each layer of \(\epsilon_\phi\). The satellite-temporal denoising process can be formulated as 
\begin{equation}
    \boldsymbol{z}'_{t-1}=DDIM(\boldsymbol{z}'_{t},\epsilon_\phi(\boldsymbol{z}'_t,c_{\phi},t),t)
\end{equation}

During training, at each time step \(\boldsymbol{t}\), the model progressively applies Gaussian noise \(\epsilon \sim \mathcal{N}(0,1)\) to the previous latent feature \(\boldsymbol{z'}_{t-1}\) to yield a new noisy feature \(\boldsymbol{z'}_t\) and learns to predict the noise by minimizing the mean-squared error:
\begin{equation}
    \mathcal{L} = \mathbb{E}_{\boldsymbol{z}'_0,t,\boldsymbol{c}_{\phi}, \epsilon \sim \mathcal{N}(0,1)} \left[ \left\| \epsilon - \epsilon_{\phi}(\boldsymbol{z}'_t, t, \boldsymbol{c}_{\phi}) \right\|_2^2 \right]
\end{equation}

\subsection{Sat2GroundScape dataset}
\label{sec:dataset}
Most existing satellite-to-ground datasets \cite{workman2015localize, zhu2021vigor, liu2019lending, xu2024geospecific} include only sparse ground collections, which limits advancements in ground video or ground scene generation. We expand this task by generating multiple ground views that are available in both panoramic and perspective formats, where some examples are shown in \cref{fig:dataset}. \textbf{Satellite Data.} We use publicly available multi-view satellite data from the 2019 Data Fusion Contest \cite{lian2020large}, covering Jacksonville, Florida. Following GVG \cite{xu2024geospecific}, we reconstruct a 3D model from the satellite views using a stereo matching method \cite{hirschmuller2007stereo}, and calculate the appearance by applying texture mapping \cite{ling2023large} from the satellite views onto the 3D model. \textbf{Ground Data.} Ground images are collected from Google Street View, with the interval range from 3 to 10 meters. Each image is panoramic and includes geolocation data (longitude, latitude, elevation) as well as orientation information (heading, pitch, roll). \textbf{Data Alignment.} Although both satellite and ground data are geo-referenced, systematic errors in gravity direction still exist and must be corrected. We manually adjust the satellite 3D model along the gravity direction to align building outlines in rendered satellite views with those in ground truth views. \textbf{Dataset Generation.} With the aligned satellite 3D model and densely sampled ground camera poses, we render appearances and depth maps in the panoramic format using Blender \cite{blender}. Perspective images are then resampled from the panoramas using predefined camera settings. Compared to GVG, which contains 7,000 pairs, we have created a significantly denser dataset with 25,000 satellite-ground pairs in panoramic format and over 100,000 pairs in perspective format. See supplementary material for detailed information on the dataset and processing methods.

\begin{figure*}
  \captionsetup[subfigure]{justification=centering}
  \begin{subfigure}[t]{0.345\textwidth}
  \includegraphics[width=\linewidth]{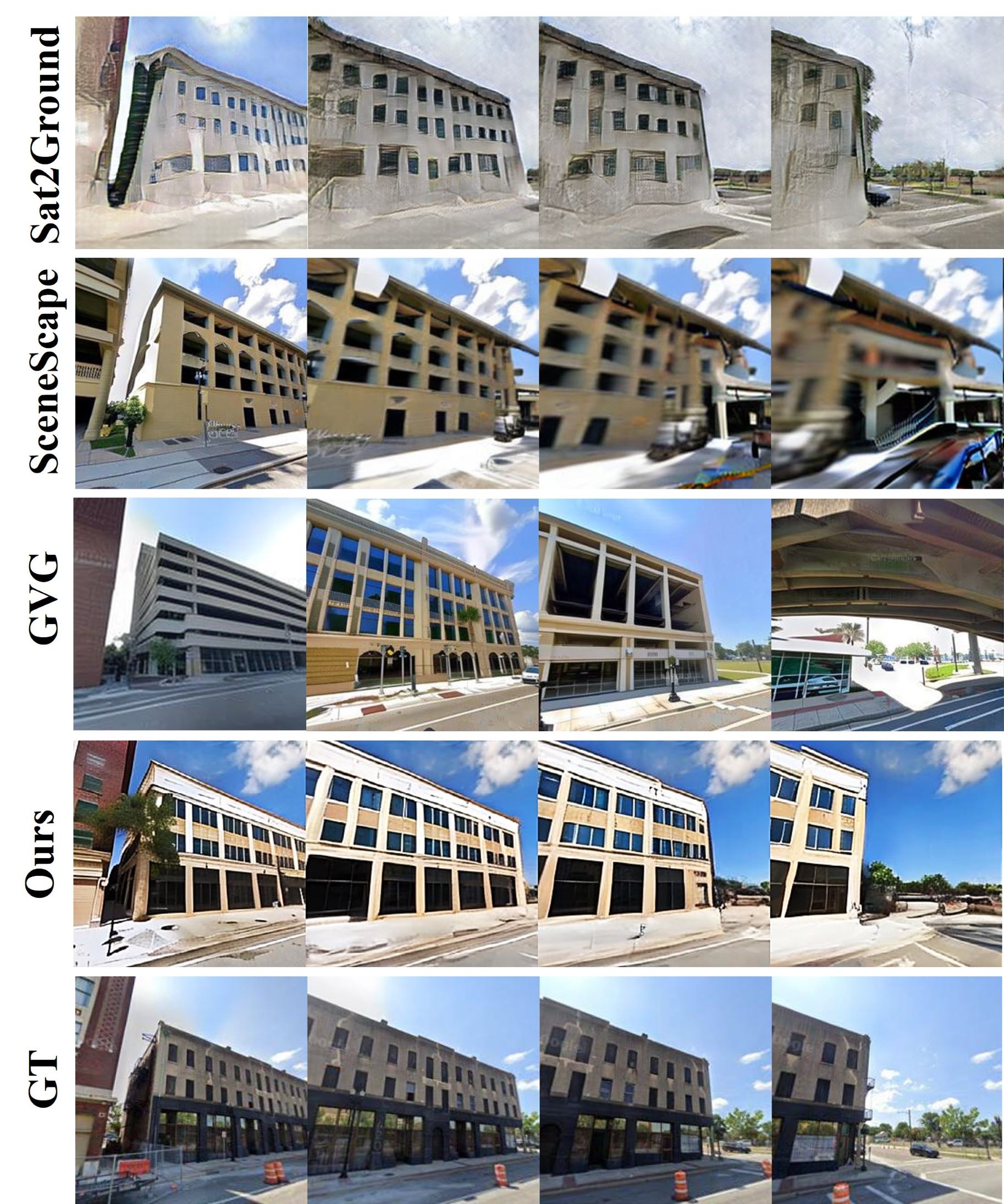}
  \end{subfigure}
    \begin{subfigure}[t]{0.317\textwidth}
  \includegraphics[width=\linewidth]{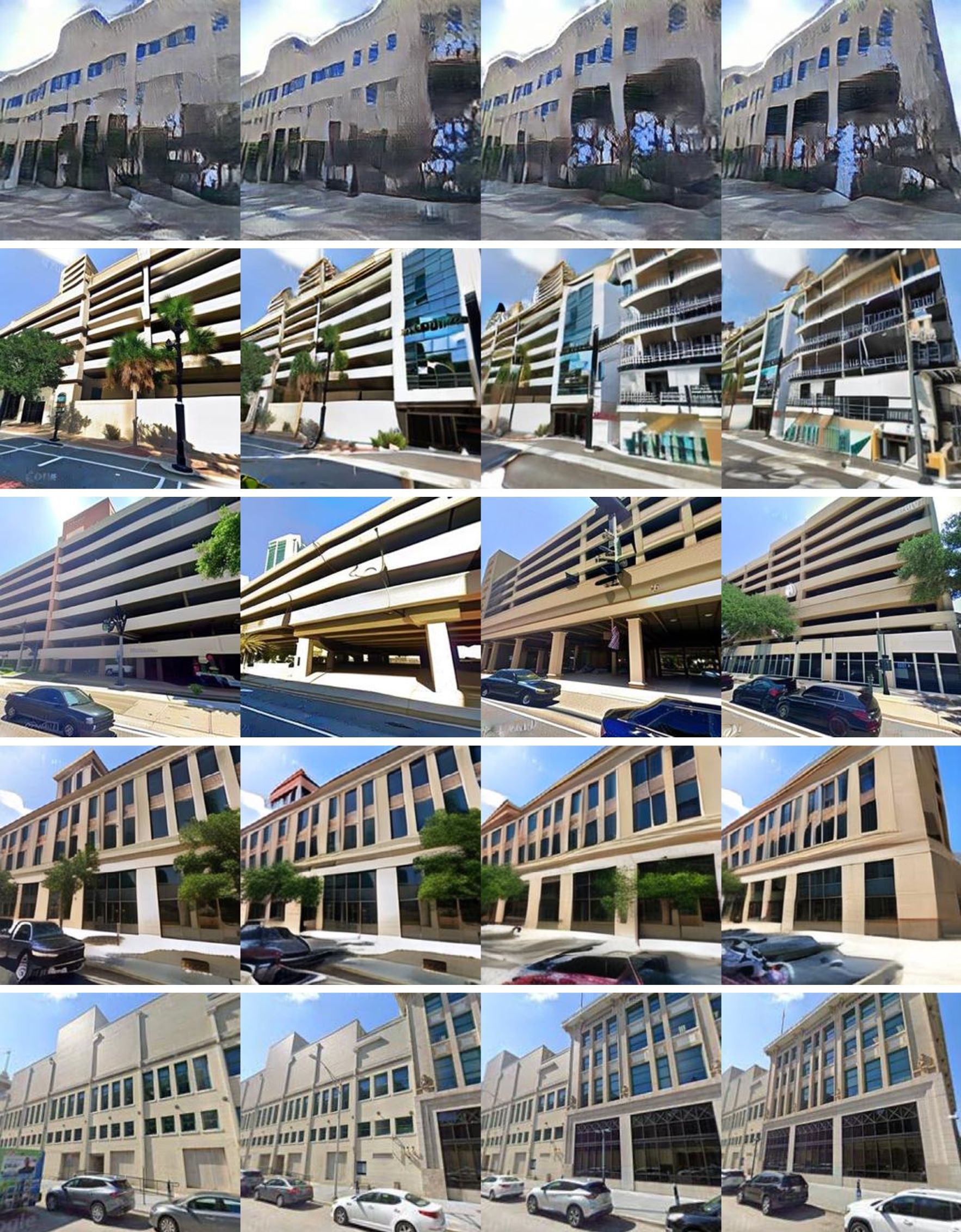}
  \end{subfigure}
    \begin{subfigure}[t]{0.317\textwidth}
  \includegraphics[width=\linewidth]{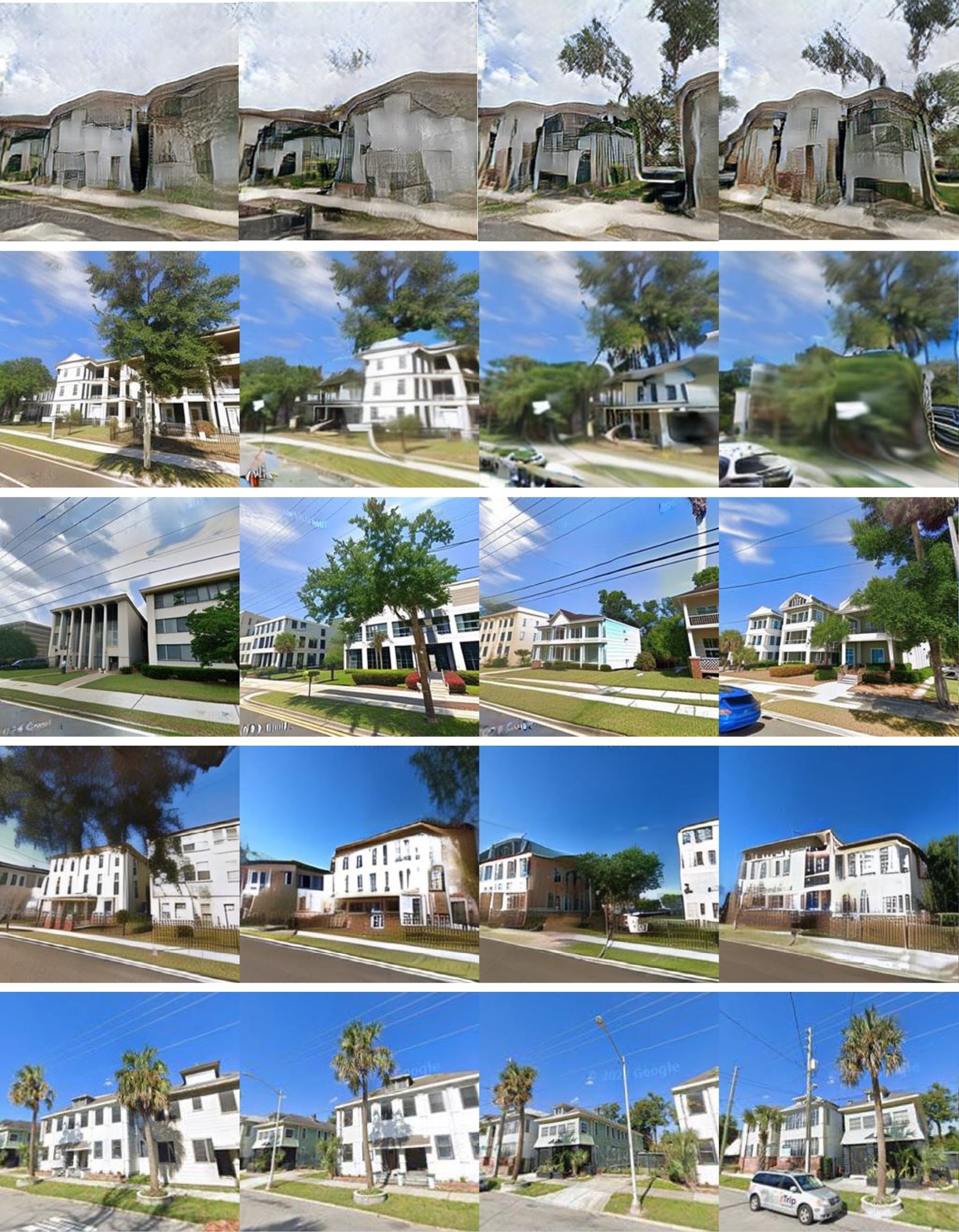}
  \end{subfigure}
\caption{\textbf{Qualitative baseline comparison on the Sat2GroundScape dataset}. We present four-view outputs of our method alongside results from Sat2Ground \cite{lu2020geometry}, SceneScape \cite{Scenescape}, and GVG \cite{xu2024geospecific}. Our method consistently produces more photorealistic results than the baseline approaches. Additional results are provided in the supplementary materials.}
\label{fig:sota_quali}
\end{figure*}

\begin{table*}[tb]
  \centering
\begin{tabular}{c c c c c c c c}
    \toprule
    \textbf{Method} & \multicolumn{2}{c}{Low level} & \multicolumn{3}{c}{Perceptual Level} & \multicolumn{2}{c}{Temporal level} \\
  & PSNR (↑) & SSIM (↑) & LPIPS(↓) & FID(↓) & DreamSIM(↓) & FVD\(^1_{\times 100}\) (↓) & FVD\(^2_{\times 100}\) (↓)\\ 
    \midrule
    Sat2Ground \cite{lu2020geometry} & 15.67 & 0.216 & 0.592 & 276.802 & 0.613 & 22.21 & 22.27 \\ 
    SceneScape \cite{Scenescape}     & \textbf{16.32} & 0.170 & 0.621 & 210.450 & 0.740 & 19.47 & 19.53 \\ 
    GVG \cite{xu2024geospecific}     & 14.59 & 0.175 & 0.581 & 175.135 & 0.600 & 19.34 & 19.39 \\  
    Ours                             & 15.86 & \textbf{0.231} & \textbf{0.542}  & \textbf{159.636} & \textbf{0.531} & \textbf{16.83} & \textbf{16.88} \\  
    \bottomrule
  \end{tabular}
  \caption{\textbf{Quantitative baseline comparison}. Our approach surpasses the baselines in both perceptual and temporal consistency metrics. FVD\(^1_{\times 100}\) and FVD\(^2_{\times 100}\) are used to assess the similarity between image sequences, with FVD\(^1_{\times 100}\) based on StyleGAN \cite{skorokhodov2022stylegan} and FVD\(^2_{\times 100}\) based on Videogpt \cite{yan2021videogpt}.}
  \label{tab:sota_quan}
\end{table*}
\section{Experiments}
\subsection{Experimental details}
\textbf{Training.} The pretrained LDM, \(\epsilon_{\theta}\), is built on Stable Diffusion v2-1 \cite{rombach2022high}. In the satellite-guided denoising process, \(\mathcal{E}_{sat}\) adopts a ControlNet-like architecture \cite{zhang2023adding} to extract the high-level satellite layout features. Unlike GVG \cite{xu2024geospecific}, which utilizes both appearance and edge maps, we find that appearance alone provides sufficient high-frequency information. In the satellite-temporal denoising process, \(\mathcal{E}_{\phi}\) employs a simple ResNet architecture \cite{he2016deep} to effectively capture camera motion.  The model training follows the diffusion noise prediction objective from DDPM \cite{ho2020denoising}, with a learning rate of \(1 \times 10^{-5}\). The framework runs on a single NVIDIA RTX 6000 Ada with a memory of 48GB. Training \(\mathcal{E}_{sat}\) and \(\mathcal{E}_{\phi}\) separately takes approximately two days in total. 

\noindent\textbf{Baselines.} We compare our method to three baselines. 
\begin{itemize}
\item \textit{Sat2Ground} \cite{lu2020geometry} and \textit{GVG} \cite{xu2024geospecific}. These two methods represent the SOTA in satellite-to-ground synthesis. Sat2Ground is a GAN-based method for generating ground views conditioned on satellite geometry and semantic information. GVG, on the other hand, is a diffusion-based approach that conditions satellite appearance and high-frequency information. For a fair comparison with our method, both can be readily adapted to generate perspective images by altering the format of their conditioning inputs from panorama to perspective format.
\item \textit{SceneScape} \cite{Scenescape} represents a SOTA approach for long-term video generation, producing multiple views sequentially, with the generation of the current view conditioned on the previous one. It ensures multi-view consistency by wrapping the previous view to the current camera pose and inpainting any occluded regions. Although it was not originally designed for satellite-to-ground tasks, we adapted it by initializing the first view with our satellite-conditioned denoised result and generating subsequent views using its depth-conditioned model.
\end{itemize}

\noindent\textbf{Datasets.} We conduct our experiments on two datasets: our own Sat2GroundScape dataset and the publicly available HoliCity dataset \cite{zhou2020holicity}. HoliCity is a city-scale dataset covering a 1000 \(\times\) 1000 meter region in central London, and includes a CAD model as well as over 6,000 ground view panorama images. Since the dataset does not include satellite data, we collected it from online sources and applied the same processing pipeline to project the satellite appearance onto the CAD model to generate the textured mesh, as detailed in \cref{sec:dataset}. 

\noindent\textbf{Metrics.} We spatially partition our dataset into 90 non-overlapping scenes, each covering an area of 600 \(\times\) 600 meters and containing approximately 500 ground collection sequences, as illustrated in \cref{fig:dataset}. We randomly select 70 scenes for training, while the remaining 20 scenes, which contain around 10,000 sequences, are used for evaluation. For quantitative assessment, we employ standard metrics such as LPIPS \cite{zhang2018unreasonable}, FID \cite{heusel2017gans}, and Dreamsim \cite{fu2023dreamsim} to evaluate the quality of the generated images by measuring the perceptual similarity between generated and real images. Additionally, to assess multi-view consistency, we use FVD \cite{unterthiner2019fvd}, the video version of FID, which provides a more comprehensive evaluation of overall quality.

\subsection{Comparing to the state-of-the-art}
\textbf{The qualitative evaluation results} for three sites are presented in \cref{fig:sota_quali}. For all methods, we generate five consecutive views corresponding to the given satellite views (the first four views are shown for clarity), with each neighboring view separated by a 10-meter distance. Our method demonstrates the best consistency across neighboring views for all three samples. While SceneScape is designed for consistent view generation, its sequential generation mechanism leads to artifact accumulation, resulting in poor-quality views after several iterations. Furthermore, the low-resolution satellite data causes blurriness during its warping process. GVG produces photorealistic results but fails to maintain consistency across neighboring views, as each view is generated independently. In contrast, Sat2Ground generates distorted results with numerous artifacts. \textbf{The quantitative evaluation results} on the Sat2GroundScape dataset can be found in \cref{tab:sota_quan}. Our method outperforms all others across the three primary metrics, with the exception of PSNR, where we are second only to SceneScape \cite{Scenescape}. For temporal-level metrics, specifically FVD, our method achieves the best performance. SceneScape, due to its sequential generation mechanism, exhibits lower performance on both FVD and perceptual-level metrics, as previously explained.

\subsection{Ablation study}
We further conduct ablation experiments to validate the effectiveness of the two core components in our method.
\begin{itemize}
\item \textbf{w/o sat}. Instead of using the satellite-guided denoising process to generate the initial ground view, we use the standard LDM to create the initial view and rely on the satellite-temporal denoising process for generating multiple ground views.
\item \textbf{w/o temp}. The satellite-temporal denoising process is removed, and only the satellite-guided denoising process is used to generate each view individually.
\item \textbf{w/o temp-sat}. Both the satellite and temporal conditioning denoising processes are removed, leaving the standard pre-trained LDM to generate each view independently.
\end{itemize}

\begin{figure}[t]
    \centering
    \includegraphics[width=\linewidth]{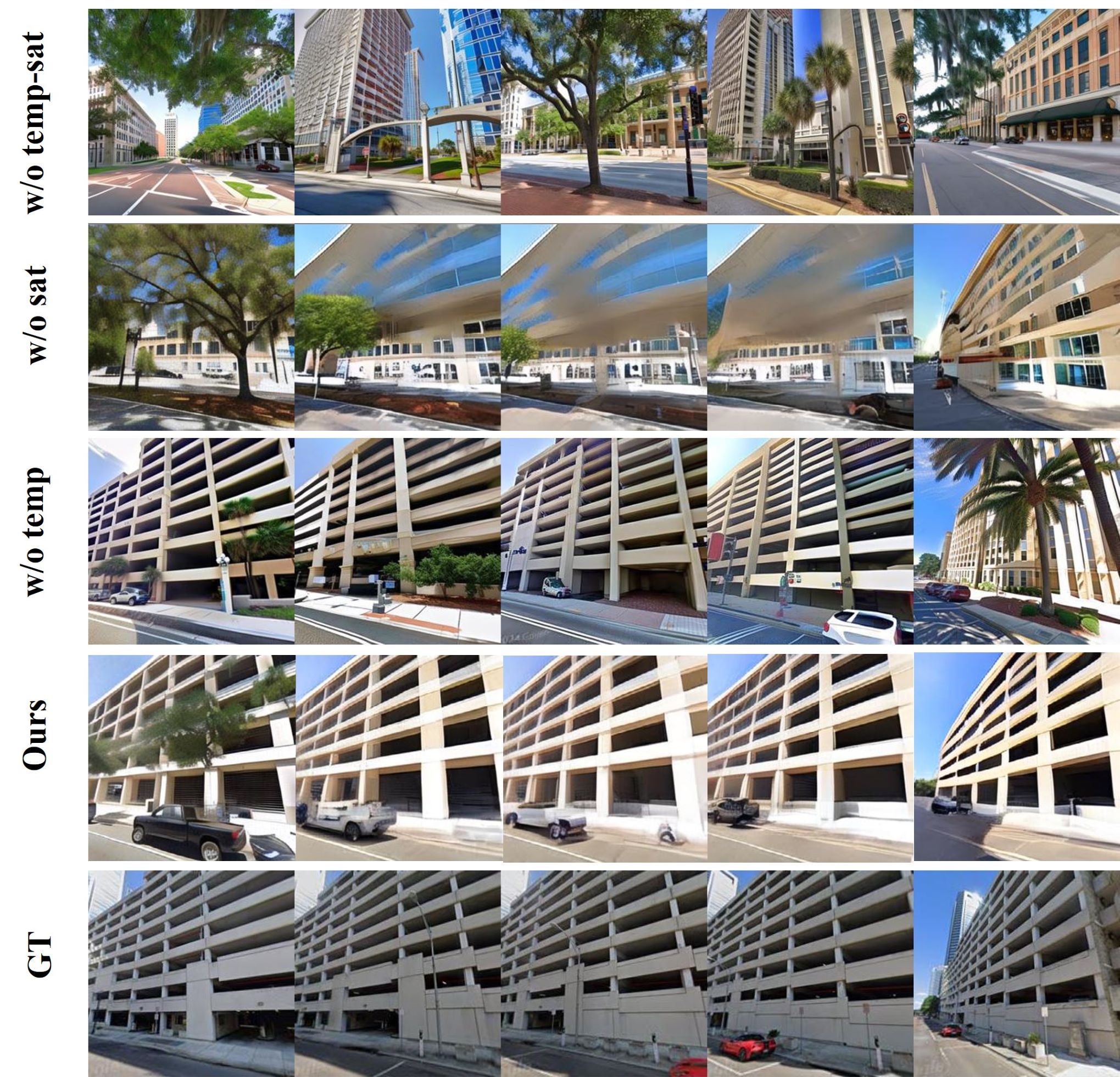}
    \caption{\textbf{Qualitative Ablation Study.} In \textbf{"w/o temp-sat"}, we show five independently generated ground views without either satellite or temporal conditioning, leading to random and unstructured outputs. In \textbf{"w/o sat"}, with a randomly generated initial view, our satellite-temporal denoising process manages to approximate the ground layout in adjacent views, demonstrating some consistency. \textbf{"w/o temp"} illustrates that while the satellite-guided denoising process alone can capture the basic ground layout, it falls short in maintaining visual coherence across neighboring views.}
    \label{fig:aba_quali}
\end{figure}

We evaluate these variants by removing each component individually from the full model, presenting qualitative results in \cref{tab:aba_quan} and quantitative results in \cref{fig:aba_quali}. Starting from the baseline “w/o temp-sat” model (i.e., the standard LDM), we observe that while photorealistic ground views are generated, they are unconditioned on satellite views, leading to meaningless outputs. Adding satellite-guided denoising (variant “w/o temp”) allows the generated views to recover the ground scenes; however, buildings and layouts are inconsistent across neighboring views. In “w/o sat,” where the first view is generated randomly and neighboring views are produced via our satellite-temporal denoising process, we see that, although the initial view lacks ground layout accuracy, subsequent views gradually recover the layout and maintain consistency across neighboring frames.

\subsection{Generalization}
\begin{figure}[t]
    \centering
    \includegraphics[width=\linewidth]{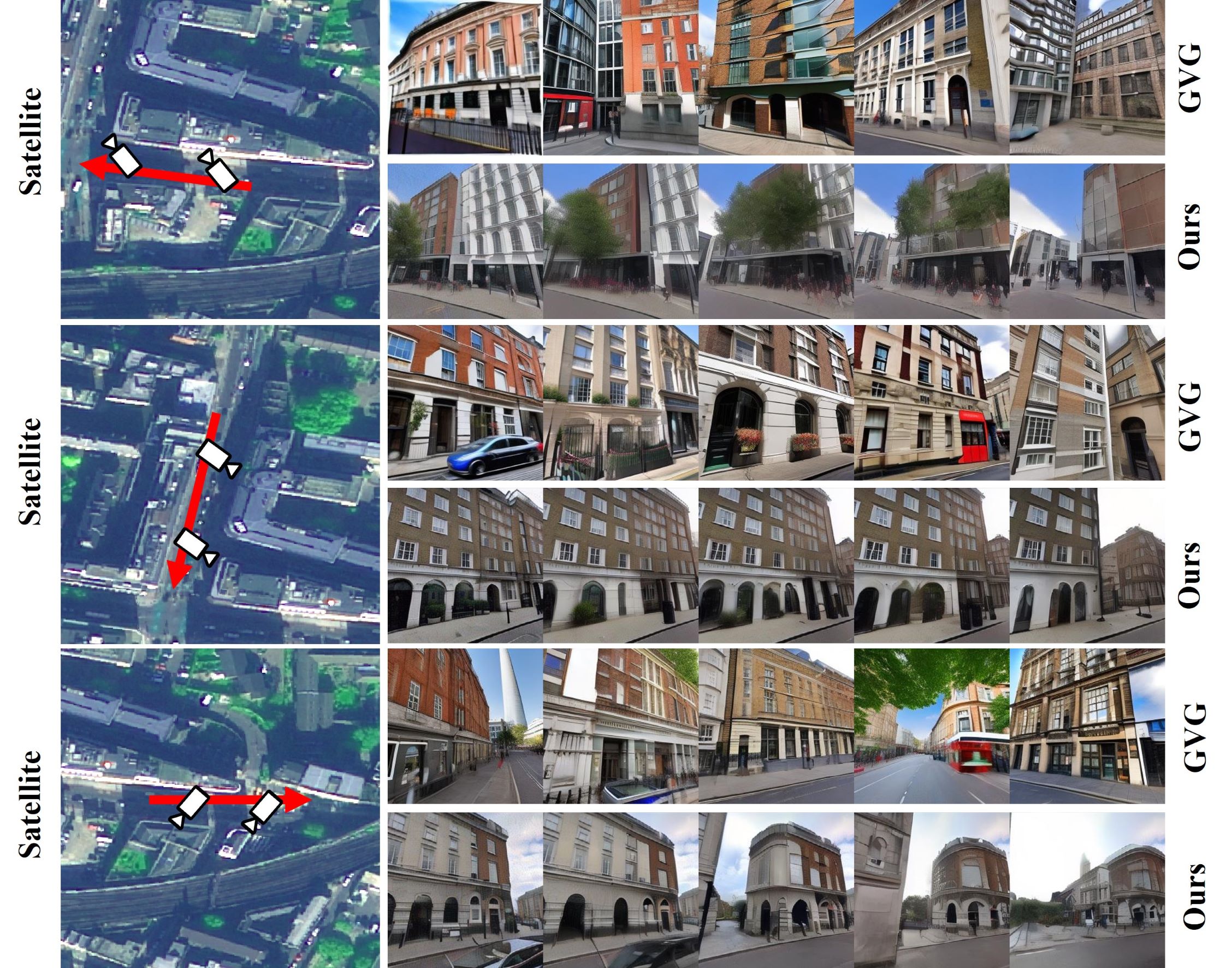}
    \caption{\textbf{Ground views generated on the Holicity \cite{zhou2020holicity} dataset.} Our method demonstrates superior generalizability and multi-view consistency compared to GVG \cite{xu2024geospecific}.}
    \label{fig:general}
\end{figure}

To demonstrate the effectiveness of our method for satellite-to-ground cross-view generation and highlight its generalization capability, we conducted additional experiments on the HoliCity dataset \cite{zhou2020holicity}. As HoliCity includes only ground-level imagery and 3D models, we collected satellite imagery from online sources and applied our approach to generate ground views. For each scene, we established ground-view navigation trajectories with a step size of 10 meters, using perspective camera settings directed toward the left-forward, right-forward, left-rear, and right-rear angles. We qualitatively compared our method with GVG \cite{xu2024geospecific} on selected scenes, as shown in \cref{fig:general}. Our approach consistently generates frames with high spatial and angular consistency across different positions and view angles. In contrast, GVG produces more variable building appearances, displaying low consistency across multiple views. 

\begin{table}
  \centering
\begin{tabular}{c c c c}
    \toprule
    \textbf{Method} & SSIM (↑) & LPIPS(↓) & FVD\(^1_{\times 100}\) (↓)\\ 
    \midrule
    w/o temp-sat & 0.106 & 0.654 & 34.83 \\ 
    w/o sat      & 0.159 & 0.630 & 22.19 \\ 
    w/o temp     & 0.176 & 0.575 & 19.21 \\  
    Ours         & \textbf{0.231} & \textbf{0.542} & \textbf{16.83} \\  
    \bottomrule
  \end{tabular}
  \caption{\textbf{Abalative evaluation of our method}. We quantitatively evaluate the influence of different components.}
  \label{tab:aba_quan}
\end{table}

\section{Conclusion}
In this paper, we present a novel framework for predicting multiple consistent ground-view images from multi-view satellite imagery. Our approach introduces a satellite-guided denoising process that guides a standard LDM to accurately generate ground views corresponding to the input satellite data. Additionally, we propose a satellite-temporal denoising process, enabling the generation of multiple consistent ground views by conditioning on both satellite data and the initially generated view. We also introduce a new satellite-to-ground dataset, supporting large-scale ground scenes and video generation from satellite imagery. Our experiments show that our method achieves a substantial performance improvement over existing baselines, producing photorealistic and consistent ground views from multi-view satellite images.

\section{Acknowledgements}
The authors are supported by the Office of Naval Research (Award No.  N000142312670) and Intelligence Advanced Research Projects Activity (IARPA) via Department of Interior/ Interior Business Center (DOI/IBC) contract number 140D0423C0034.

{
    \small
    \bibliographystyle{ieeenat_fullname}
    \bibliography{main}
}

\end{document}